\documentclass[letterpaper]{article} 
\usepackage{aaai23}  
\usepackage{times}  
\usepackage{helvet}  
\usepackage{courier}  
\usepackage[hyphens]{url}  
\usepackage{graphicx} 
\usepackage{natbib}  
\usepackage{caption} 
\usepackage{algorithm}
\usepackage{algorithmic}

\usepackage{mathtools}
\usepackage{lipsum}
\usepackage{caption, subcaption}
\usepackage{url}
\usepackage{amsmath, amsfonts, amsthm, bm}
\usepackage{soul}
\usepackage{tabularx, multirow}
\usepackage{dsfont}
\usepackage{booktabs}
\usepackage{xcolor}
\usepackage{chngcntr}

\usepackage{xspace}
\newcommand{\proposed}{TopExpert\xspace}
\newcommand{\smallsection}[1]{{\vspace{0.05in} \noindent \bf {#1.\hspace{5pt}}}}

\usepackage{newfloat}
\usepackage{listings}
\DeclareCaptionStyle{ruled}{labelfont=normalfont,labelsep=colon,strut=off} 
\floatstyle{ruled}
\newfloat{listing}{tb}{lst}{}
\floatname{listing}{Listing}
\title{Learning Topology-Specific Experts for Molecular Property Prediction}
\author{
    Suyeon Kim\textsuperscript{\rm 1},
    Dongha Lee\textsuperscript{\rm 2},
    SeongKu Kang\textsuperscript{\rm 1},
    Seonghyeon Lee\textsuperscript{\rm 1},
    Hwanjo Yu\textsuperscript{\rm 1}\thanks{Corresponding Author} 
}
\affiliations{
    \textsuperscript{\rm 1}Pohang University of Science and Technology (POSTECH), South Korea\\
    \textsuperscript{\rm 2}University of Illinois at Urbana-Champaign (UIUC), United States\\
    \{kimsu, seongku, sh0416, hwanjoyu\}@postech.ac.kr, donghal@illinois.edu
}

\begin{document}

\maketitle

\begin{abstract}
Recently, graph neural networks (GNNs) have been successfully applied to predicting molecular properties, which is one of the most classical cheminformatics tasks with various applications. 
Despite their effectiveness, we empirically observe that training a single GNN model for diverse molecules with distinct structural patterns limits its prediction performance.
In this paper, motivated by this observation, we propose \proposed to leverage topology-specific prediction models (referred to as experts), each of which is responsible for each molecular group sharing similar topological semantics.
That is, each expert learns topology-specific discriminative features while being trained with its corresponding topological group.
To tackle the key challenge of grouping molecules by their topological patterns, we introduce a clustering-based gating module that assigns an input molecule into one of the clusters and further optimizes the gating module with two different types of self-supervision: 
topological semantics induced by GNNs and molecular scaffolds, respectively. 
Extensive experiments demonstrate that \proposed has boosted the performance for molecular property prediction and also achieved better generalization for new molecules with unseen scaffolds than baselines.  
The code is available at https://github.com/kimsu55/ToxExpert.
\end{abstract}

\section{Introduction}

\begin{figure}[t]
    \centering
    \includegraphics[width=\linewidth]{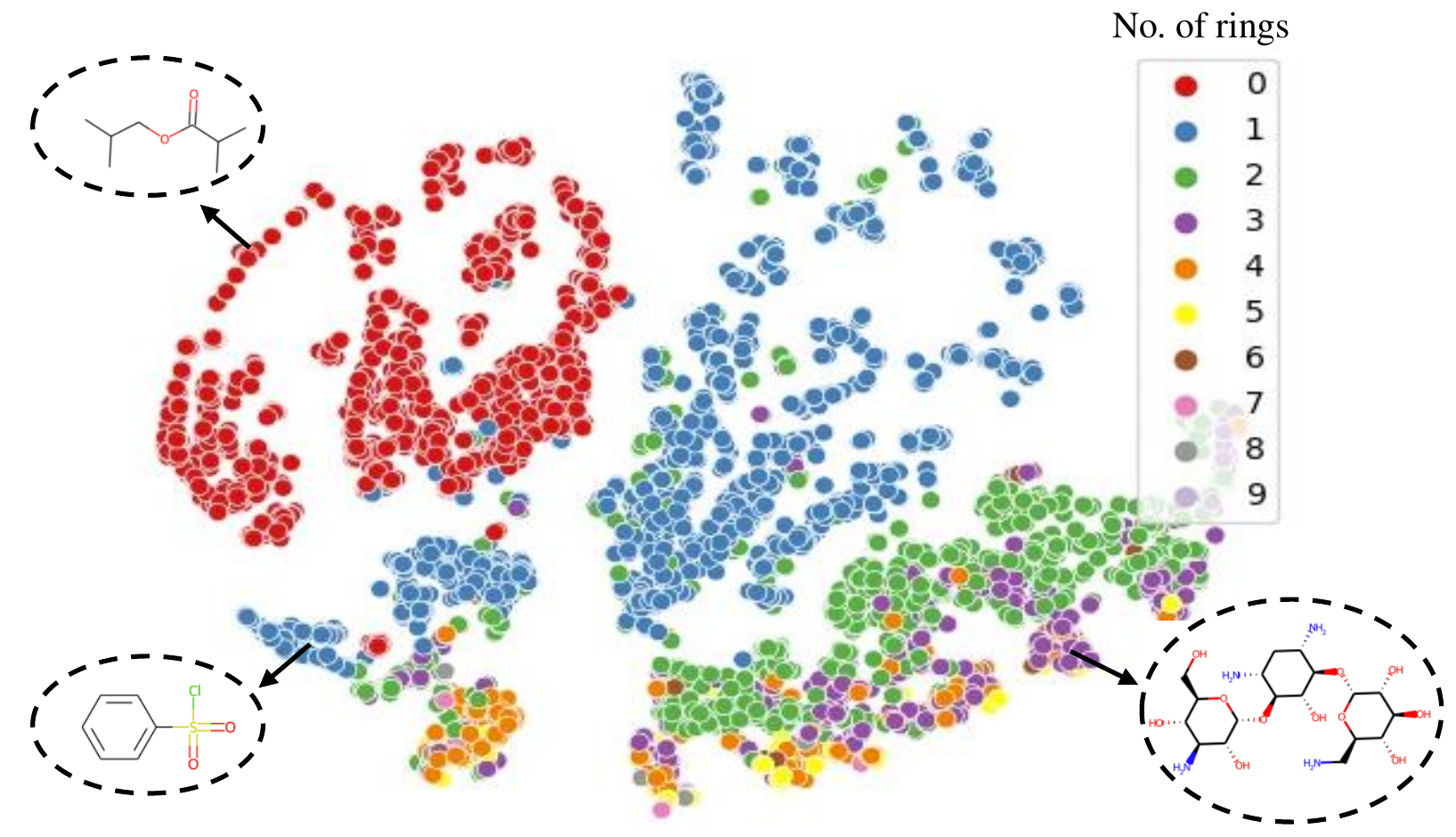}
    \caption{The t-SNE visualization of molecular representations produced by randomly-initialized GIN~\cite{xu2018powerful}.}
    \label{fig:motivation1}
\end{figure}

Molecular property prediction is one of the most classical cheminformatics tasks with applications in various areas such as quantum chemistry \cite{gilmer2017neural, brockschmidt2020gnn} and biology \cite{jumper2021highly, liu2020deep}. 
It aims to clarify the correspondence between a molecule and its properties, including toxicity, barrier permeability, and adverse drug reactions. 
Molecular topology, which refers to the pattern of interconnections among atoms and thus determines the ultimate structure of a molecule \cite{rouvray1986predicting}, has played an essential role in predicting the properties. 
It is well known that molecules with similar structural patterns are likely to have similar properties \cite{lo2016chemical}, and accordingly, such topological information has been widely exploited for drug discovery and further modifications to improve target properties \cite{hu2016computational, naveja2019systematic}.

Recently, Graph neural networks (GNNs) have been successfully applied to predicting molecular properties \cite{yang2019analyzing, yang2020factorizable, yu2022molecular}. Specifically, regarding molecules as graphs, GNNs extract localized spatial features by aggregating information at neighbor nodes and iteratively combining them to construct high-level representations, which effectively encodes the molecular topology (Figure \ref{fig:motivation1}). 
Several attempts have been made to help GNNs further capture topology information. 
One approach is to leverage frequently-occurred subgraphs, known as motifs, which are considered as basic meaning blocks with particular functions \cite{fey2020hierarchical, peng2020motif}. 
Another approach is to preserve structural information in the process of aggregating node representations into a graph level \cite{baek2021accurate,lee2021learnable}.

\begin{figure}[t]
    \centering
    \includegraphics[scale=0.5]{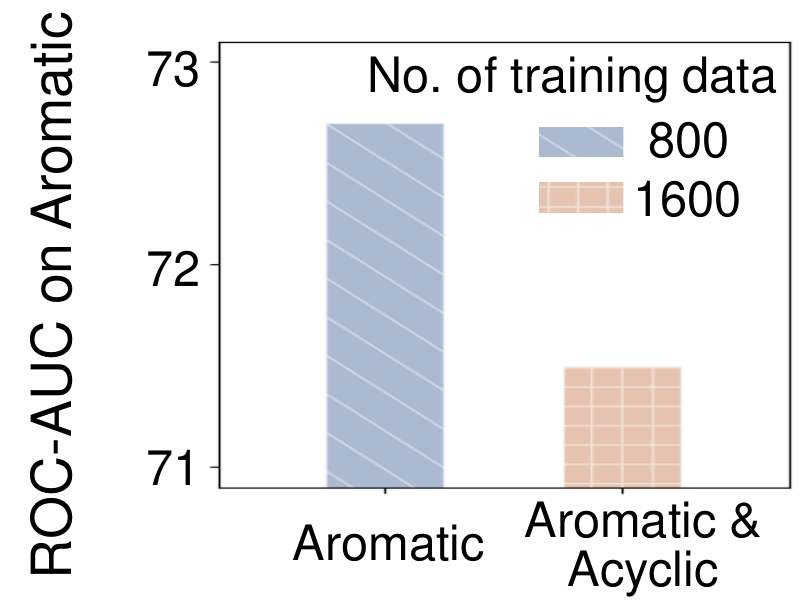}
    \caption{Classification performance on aromatic molecules.}
    \label{fig:motivation2}
\end{figure}

Despite their effectiveness, we observe that training a single GNN model to learn molecules with distinct structural patterns can hinder the model from capturing topology-specific features and eventually limits its performance. 
In Figure \ref{fig:motivation2}, for the property prediction of aromatic molecules, one GNN model trained on both aromatic and acyclic compounds (i.e., molecules having distinct topology) performs worse than another GNN model trained on aromatic compounds only despite its larger number of training data.
In this sense, we argue that there is much room for improvement in case of utilizing topology-specialized models (called \textit{experts}) for each molecule group sharing similar topological semantics, instead of a single model.

One simple solution for grouping molecules by their topology is using \textit{molecular scaffolds}. 
A scaffold is generally known as a core structure of a molecule, and each molecule is deterministically assigned to one of the scaffolds based on pre-defined rules.
In particular, it has been well studied and exploited in pharmacological and biological areas \cite{hu2016computational}. 
However, there are two technical challenges to utilizing molecular scaffolds. 
First, scaffolds are too fine-grained to be directly used.
For instance, there exist 7,831 molecules with 2,405 distinct scaffolds in the Tox21 dataset. 
Thus, assigning an expert to each scaffold is not feasible, as the number of molecules in each scaffold is too few to train an expert;
only 3.3 molecules share the same scaffold on average for Tox21 dataset. 
Second, at the deployment of the model, it will face the challenge of handling new molecules of \textit{unseen} scaffolds that are not in the training set. 
This arises from the scaffold splitting procedure, which enforces training and testing datasets sharing no common scaffolds;
this has been widely used as a default evaluation setup because it provides a more realistic estimate of the model performance than random splitting~\cite{wu2018moleculenet}. 
In this regard, simply using scaffold indices as the input features is not applicable because unseen scaffolds cannot be handled during testing.

\begin{figure}[t]
    \centering
    \includegraphics[width=\linewidth]{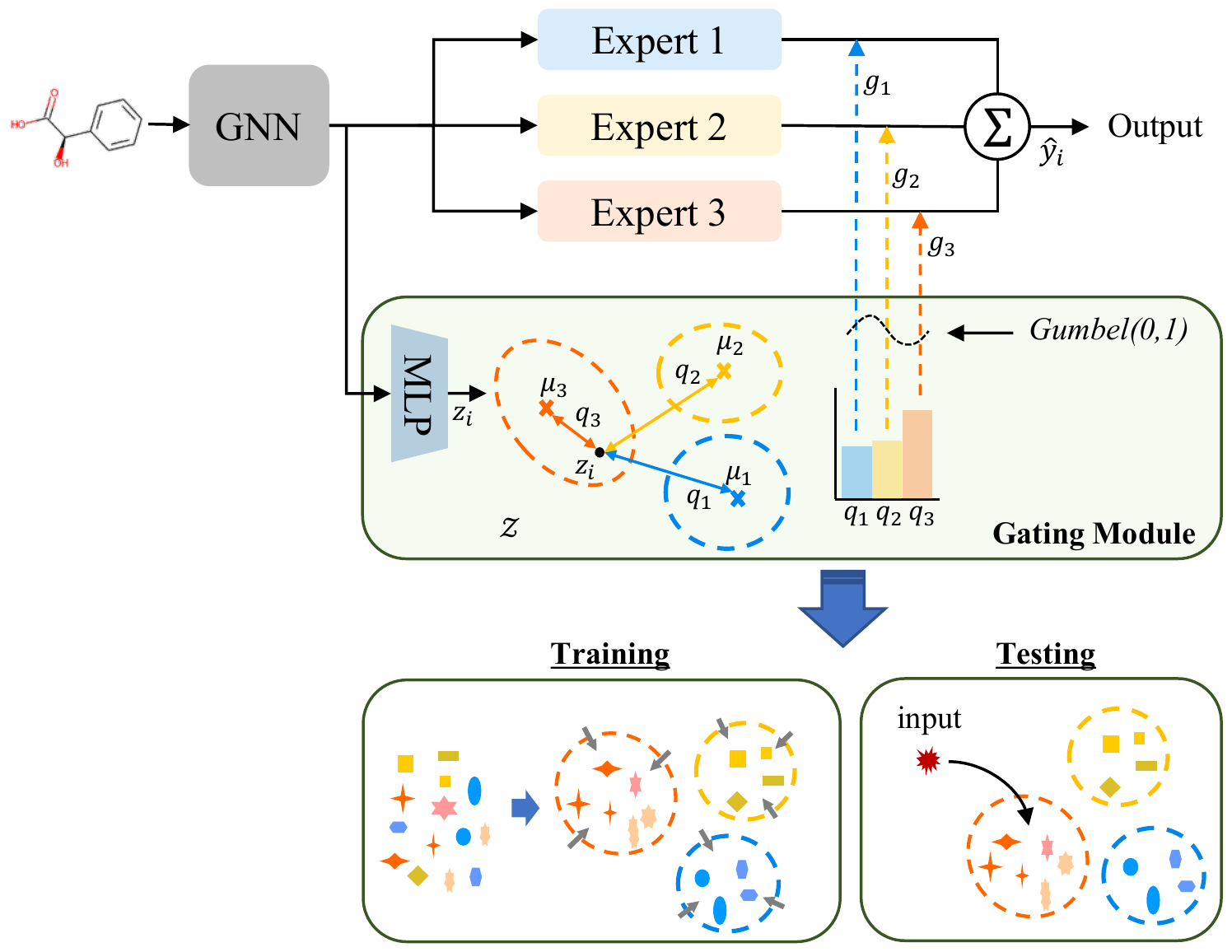}
    \caption{The overall framework of \proposed. It consists of three components: GNN, a gating module, and multiple experts. In training, the gating module optimizes molecular representations to be distinguishable according to their topological information. In testing, it selects experts more relevant to an input molecule with a higher weight.}
    \label{fig:arch}
\end{figure}

In this paper, we propose a topology-specific expert model, named \textbf{\proposed}, that exploits multiple experts specialized in their corresponding topological groups (Figure~\ref{fig:arch}).
Each expert is responsible for predicting the property of molecules in each group by effectively capturing discriminative features specific to its associated topological group;
it serves as topology-specific knowledge.
For the assignment of molecules to the experts, we introduce a novel gating module that identifies topological groups relevant to an input molecule based on its cluster assignment.
  
To obtain effective clusters of molecules based on their topology, we optimize our gating module in a self-supervised manner, in order to make molecular representations distinguishable according to their topological semantics. 
By using both topological features (encoded by a GNN) and prior knowledge (induced by scaffolds) of molecules, 
we strengthen (1) the cohesion of each cluster and (2) the alignment between fine-grained scaffolds and coarse-grained clusters.
In the end, \proposed clusters molecules having similar topological semantics into the same group, which allows us to identify relevant groups even for a new molecule with an unseen scaffold.
The contribution of this work is threefold as follows:    
\begin{itemize}
    \item We demonstrate that learning molecules having distinct structural patterns with a single model may negatively affect capturing topology-specific features for property prediction, which has not been studied in previous works. 

    \item We propose topology-specific experts with a novel clustering-based gating module that leverages topological information induced by GNNs and molecular scaffolds.  
    To the best of our knowledge, this is the first work to incorporate molecular scaffolds into GNNs.   
    
    \item We validate the effectiveness of \proposed as a general tool for boosting the performance of existing GNN models through extensive experiments.  
    Furthermore, we show that our model can be well-generalized to new molecules with unseen scaffolds.

\end{itemize}

\section{Related Work}
\smallsection{Molecular Property Prediction} 
It aims to clarify the correspondence between a molecule and its properties.
For encoding molecules as numerical features, traditional approaches use hand-engineered molecular features such as molecular fingerprints \cite{rogers2010extended} and molecular descriptors \cite{hansen2015machine}. 
However, such hand-crafted features are insufficient to reflect structural similarities and biological activities of molecules. 
Recently, GNNs have gained significant attention because of their convincing performance \cite{gilmer2017neural, lu2019molecular, klicpera2020directional}. 
GNNs also have the advantage that they can be directly applied to non-euclidean graph data without feature engineering. 
To better exploit molecular topology, several studies have leveraged motifs (i.e., frequently-occurred subgraphs) mostly obtained based on prior knowledge \cite{fey2020hierarchical, yu2022molecular}. 
In addition, some attempts have been made to better preserve structural information in the process of aggregating node-level representations into a graph-level one \cite{baek2021accurate,lee2021learnable}.
Another line of work has focused on pre-training and self-supervised learning of GNNs to improve the performance on downstream tasks \cite{sun2019infograph, hu2020strategies, subramonian2021motif}.

Although the aforementioned methods have shown promising results with the help of their enhanced GNNs, they mainly utilize a single GNN model for their target task.
However, our empirical findings imply that such a single model is insufficient to handle molecules with distinct structural patterns.
For this reason, this work focuses on utilizing multiple experts to capture topological group-specific knowledge for property prediction, which is orthogonal to the existing studies.

\smallsection{Expert Models} 
It mainly employs a divide-and-conquer strategy for systems composed of many distinct experts, each of which learns to handle a part of input space \cite{jacobs1991adaptive}.  
Recently, this idea of mixture of experts (MoE) has been combined with neural networks, and
\citet{shazeer2017outrageously} firstly propose a sparsely-gated MoE with the goal of reducing computational costs.
In particular, MoE architecture has been validated in various applications such as image recognition \cite{ahmed2016network}, machine translation \cite{shen2019mixture}, and recommender system \cite{kang2020rrd, kang2021topology}. 
When it comes to graph classification, \citet{hu2021graph} adopts a multi-gate MoE \cite{ma2018modeling} to tackle the performance deterioration caused by class-imbalance.
However, they do not consider the explicit graph topology in the gating process, and its performance is degraded when using a single gating network \cite{hu2021graph}.
Our work is distinguished from the above approaches in that we elaborate MoE to leverage topology knowledge induced by GNNs and molecular scaffolds in the gating process.

\section{Preliminaries}

\subsection{Problem Formulation} 

In molecular property prediction, a molecule is usually regarded as a graph whose nodes and edges respectively correspond to the atoms and chemical bonds in the molecule. 
Given a training set $D=\{(G_i,y_i)\}^{N}_{i=1}$, where $G_i \in \mathcal{G}$ denotes a graph (i.e., molecule) and $\mathrm{y}_i \in \{0,1\}^T$ is its binary labels (i.e., target properties) for $T$ tasks, the goal of our work is to train a model $f:\mathcal{G}\to \mathcal{Y}$ to predict properties of new input molecules.
A graph $G$ with $n$ nodes and $m$ edges is defined by its node features $X\in \mathbb{R}^{n\times d_{node}}$, edge features $E\in \mathbb{R}^{m\times d_{edge}}$, and the adjacency matrix $A\in \mathbb{R}^{n\times n}$, where $A(u,v)=1$ if there is an edge between node $u$ and node $v$, and $A(u,v)=0$ otherwise.

\subsection{Molecular Scaffolds} 

A scaffold is generally known as a core structure of a molecule \cite{hu2016computational}, and ``Bemis and Murcko (BM) scaffolds'' whose major building blocks are ring systems with linkers have been widely used \cite{bemis1996properties}.\footnote{In this paper, we refer to BM scaffolds as scaffolds for brevity.}
Note that the scaffold of a molecule is easily determined based on pre-defined rules. 
Scaffold-inducing structural information is instructive when exploiting a direct structure-property relationship in pharmacological and biological areas. 
For example, it was reported that 5,120 known drugs were categorized into 1,179 scaffolds, and 50 percent of the total drug molecules belong to only 32 of 1,179 \cite{bemis1996properties}. 
In this sense, benchmark datasets for molecular property prediction are mainly split into training and testing sets to have disjoint scaffold indices, which makes the prediction more challenging yet realistic compared to random split \cite{wu2018moleculenet}.

\subsection{Graph Neural Networks} 
To encode useful information from graph-structured data, 
GNNs iteratively compute node representations by aggregating messages from neighbor nodes and connected edges in a layer-wise manner \cite{scarselli2008graph}.
The representation of node $u$ at the ($l$+1)-th GNN layer (denoted by $\mathrm{h}_u^{(l+1)}$) is defined by
\begin{equation}
\label{eq:gnn}
    \mathrm{h}_u^{(l+1)}=f_u^{(l)}\left(\mathrm{h}_u^{(l)}, f_a^{(l)}\left(\{(\mathrm{h}_v^{(l)}, \mathrm{h}_u^{(l)}, \mathrm{e}_{uv} ):\forall{v}\in \mathcal{N}(u)\}\right)\right),
\end{equation}
where $\mathcal{N}(u)$ is the set of neighbors of node $u$, and $f_u^{(l)}$ and $f_a^{(l)}$ are the arbitrary differentiable functions for updating representations and aggregating messages, respectively. 
In the end, the graph-level representation $\mathrm{h}_G$ is obtained by summarizing all node representations using a readout function $f_r$ such as averaging, summation, or sophisticated methods, 
\begin{equation}
\label{eq:readout}
    \mathrm{h}_G= f_r\left(\{\mathrm{h}_u^{l}|u \in G\}\right).
\end{equation}

\section{\proposed: The Proposed Method}

In this section, we present \proposed, a GNN-based framework that adopts multiple experts for molecular property prediction.
Each expert learns from molecules in the corresponding topological group, where the molecules are assigned by a clustering-based gating module.

\subsection{Overview}
The goal of \proposed is to effectively capture topology-specific features by exploiting multiple experts specialized in their corresponding topological groups.
In Figure \ref{fig:arch}, \proposed consists of three parts: a GNN, $K$ number of experts, and a gating module.
Serving as a feature extractor, the GNN maps an input molecule into a low-dimensional molecule representation.
Then, each expert, implemented as a fully-connected (FC) layer, computes its output (i.e., classification logits) by using the molecule representation.
In the end, the final output is obtained by aggregating the experts' predictions with their weight values, which are calculated by the gating module to assign the input molecule into the topological groups (i.e., molecule clusters).
To encourage the gating module to effectively assign each molecule into the clusters based on topological similarity, 
we adopt two learning objectives to make the molecule representations discriminative in terms of topological patterns: 
(1) The \textit{clustering loss} strengthens the cohesion of the clusters by increasing the maximum cluster membership of each molecule,
and (2) the \textit{alignment loss} further aligns the coarse-grained clusters with the fine-grained scaffolds to encode topological semantics induced by molecular scaffolds.

\subsection{Multiple Experts Learning for Property Prediction}

The key idea of our gating module is to assign each molecule into the experts based on its similarity to the corresponding molecule clusters.
To this end, given an input molecule $G_i$, the gating module non-linearly transforms its GNN-induced molecule representation $\mathrm{h}_{G_i}$ into the topology representation $\mathrm{z}_i\in\mathcal{Z}$, which is used for molecule clustering based on topological patterns;  
we simply use a multi-layer perceptron (MLP) with two hidden layers: $\mathrm{z}_i=\mathrm{MLP}(\mathrm{h}_{G_i}) \in \mathbb{R}^{d_\mathcal{Z}}$. 

Then, to calculate the cluster assignment probability, we introduce trainable parameters for $K$ cluster centroids, $M=[\mu_1;...;\mu_K]\in\mathbb{R}^{K\times d_\mathcal{Z}}$.\footnote{The centroids are initialized by $K$-means clustering on topology representations from a randomly-initialized model.}
To compute the assignment probability of the $i$-th molecule to the $k$-th topological cluster, denoted by $q_{ik}$, we measure the similarity between the topology representation $\mathrm{z}_i$ and the $k$-th centroid $\mu_k$ using a Student's $t$-distribution with one degree of freedom as a kernel \cite{van2008visualizing}:       
\begin{equation}
\label{eq:cluster} 
    q_{ik}={\frac{{(1+||\mathrm{z}_i-\mathrm{\mu}_k||^2)^{-1}}}{\sum_{k'}{(1+||\mathrm{z}_i-\mathrm{\mu}_{k'}||^2)^{-1}}}}.
\end{equation}

Based on the cluster assignment probability, we additionally adopt a stochastic selection strategy to help the experts to learn comprehensive topological patterns in the early training stage.
Precisely, we use the Gumbel-Softmax, \cite{jang2017categorical} which approximates the sampling from the cluster assignment distribution with the reparameterization trick for its differentiation:    
\begin{equation}
\label{eq:gumbel}
    g_{ik}={\frac{\exp{((\log{q_{ik}+\gamma_{ik}})/ \tau)}}{\sum_{k'} \exp{((\log{q_{ik'}+\gamma_{ik'}})/ \tau)}}},
\end{equation}
where $\gamma_{ik}$ is drawn from the standard Gumbel distribution; i.e., $\gamma_{ik}=-\log(-\log(U_{ik}))$ with $U_{ik} \sim Uniform(0,1)$.
The temperature of the softmax $\tau$ is initially set to the value $T_0$ and exponentially annealed down to the last value $T_E$; 
i.e., $\tau = T_0({\frac{T_E}{T_0}})^{{\frac{e}{E}}} $ such that $T_0 \gg T_E $ with the total training epochs $E$ and the current epoch $e$. 
The higher $\tau$ value makes the Gumbel-Softmax distribution more uniform over the experts \cite{jang2017categorical}, which allows experts to stochastically explore diverse topological patterns in the early training phase \cite{balin2019concrete}.

Finally, each of the $K$ experts outputs the logit $\mathrm{o}_{ik}\in\mathbb{R}^T$, and the final prediction output $\mathrm{\hat{y}}_i\in\mathbb{R}^T$ is obtained by
\begin{equation}
\label{eq:gating}
    \mathrm{\hat{y}}_i = \sum_{k=1}^{K}g_{ik}\cdot \sigma(\mathrm{o}_{ik}),
\end{equation}
where $\sigma(\cdot)$ is a sigmoid function. 
To optimize the model parameters for the target prediction tasks (i.e., multi-label binary classification), we define the binary cross entropy loss:   
\begin{equation}
\label{eq:classification}
    \mathcal{L}_{class} = {\frac{1}{N}} \sum_{i=1}^{N} \mathrm{BCE}(\mathrm{y}_i, \mathrm{\hat{y}}_i).
\end{equation} 
Note that, as training progresses, the weight distribution over all the experts, i.e., $g_{i}$, is getting closer to one-hot distribution due to the annealed temperature $\tau$ as in Eq.~\eqref{eq:gating}. 
In the end, each expert is only updated by gradients from its corresponding topology group, while gradients from the other groups hardly flow back to the expert with negligible weights. This encourages each expert to learn discriminative features from the corresponding subset of the data.

The remaining challenge is guiding the gating module to discover topological clusters of molecules, which should be effective for the target prediction tasks. 
Since there do not exist guidance labels for clustering, we optimize the topology representations and cluster centroids in a self-supervised manner by leveraging topological features (encoded by a GNN) and prior knowledge (induced by scaffolds).

\subsection{Gating Module Learning for Cohesive Clustering}
For the guidance of the gating module in terms of clustering, \proposed directly optimizes the cluster assignment distribution $\mathrm{q}_i$ (Eq.~\eqref{eq:cluster}) with the help of the target distribution $\mathrm{p}_i$ computed using the current cluster assignment.
As shown in Figure \ref{fig:motivation1}, GNNs are inherently capable of extracting discriminative structural features to some extent. 
Therefore, we define a target distribution for each molecule based on the cluster
assignment of the current model, which therefore enhances the high-confident assignment~\cite{xie2016unsupervised} via a squaring operation:
\begin{equation}
\label{eq:targetprob}
    p_{ik}={\frac{{q_{ik}^2/\sum_{i}q_{ik}}}{\sum_{k'}(q_{ik'}^{2}/\sum_{i}q_{ik'})}}.
\end{equation}
Note that this target distribution strengthens the cluster cohesion, which tunes topology representations further discriminative with respect to their topological patterns.

To sum up, we define the clustering loss that enhances the cohesion of the clusters by minimizing KL divergence between a cluster assignment distribution $\mathrm{q}_{i}$ and a target distribution $\mathrm{p}_{i}$ as follows:
\begin{equation}
\label{eq:cluster_loss}
    \mathcal{L}_{cluster} = \mathrm{KL}(P||Q)={\frac{1}{N}} \sum_{i=1}^{N}\sum_{k=1}^{K}p_{ik}\log{\frac{p_{ik}}{q_{ik}}}.
\end{equation}
    
\subsection{Gating Module Learning for Scaffold Alignment}

To incorporate rich topological information into molecule clustering, we take advantage of molecular scaffolds as prior knowledge to guide the clustering process.
Since scaffolds are much finer-grained than molecular clusters of our gating module, we propose a scaffold alignment scheme that aligns each molecule's scaffold index with its cluster assignment distribution.
This facilitates the densely-located topology representations of molecules having not only the same scaffolds but also topologically similar scaffolds.  

Inspired by the optimal transport (OT) problem \cite{monge1781memoire}, we devise a distance measure between the cluster assignment distribution $\mathrm{q}_i$ and the scaffold distribution $\mathrm{s}_i$ for a molecule $G_i$. 
Note that OT has been widely used to measure the distance between two probability measures by optimizing all possible probabilistic couplings between them \cite{ma2021unsupervised, lee2022toward}.
In the case of a training set with $V$ scaffolds, the scaffold index of each molecule $\mathrm{s}\in\{0,1\}^V$ is represented as a one-hot distribution over all the scaffolds. 
To formulate our OT problem, we define the cost matrix $\mathrm{M}\in\mathbb{R}^{V\times K}$ between $V$ scaffolds and $K$ clusters based on the cosine distance of their representations;
that is, $m_{vk}=1-\mathbf{cos}(\mathrm{e}_v, \mathrm{\mu}_k)$,
where $\mathrm{e}_v \in \mathbb{R}^{d_\mathcal{Z}} $ is the scaffold representation with trainable parameters. 
Using the cost matrix $\mathrm{M}$ with two distributions $\mathrm{s}$ and $\mathrm{q}$, our distance is defined by
\begin{equation}
\label{eq:op_dist}
    d_M(\mathrm{s,q})\coloneqq \min_{\mathrm{T}\in \mathrm{U}(\mathrm{s},\mathrm{q})}   \langle\mathrm{T},\mathrm{M}\rangle.
\end{equation}
$\langle\cdot,\cdot\rangle$ is the Frobenius dot-product and $\mathrm{T}\in \mathbb{R}^{V\times K}$ is one of the feasible transport matrices in the space $\mathrm{U(s,q)}\coloneqq\{\mathrm{T}\in (\mathbb{R^+})^{V \times K}|\mathrm{T}\mathbf{1}_{K}=\mathrm{s}, \mathrm{T}^\mathrm{T} \mathbf{1}_{V}=\mathrm{q} \}$, where $\mathbf{1}_d$ is the $d$-dimensional one vector. 
Since a scaffold distribution $\mathrm{s}$ is a one-hot distribution, Eq.~\eqref{eq:op_dist} has the closed-form solution for the transport matrix without requiring any optimization process such as Sinkhorn algorithm \cite{cuturi2013sinkhorn}; i.e., $\mathrm{T}^* = \mathrm{sq}^\mathrm{T}$.
In the end, we define the alignment loss to minimize the distance $d_M(\mathrm{s,q})$, which can be rewritten as follows:
\begin{equation}
\label{eq:align_loss}
    \mathcal{L}_{align}={\frac{1}{N}} \sum_{i=1}^{N}\sum_{k=1}^{K} \sum_{v=1}^{V} s_{iv}\cdot q_{ik}\cdot(1-\mathbf{cos}(\mathrm{e}_v, \mathrm{\mu}_k)).
\end{equation}
Minimizing the distance for a molecule $G_i$ has the effect of matching the molecule's topology representation $\mathrm{z}_i$ with its corresponding scaffold representation $\mathrm{e}_v$ while both representations approach the centroid of a specific cluster. 
When it comes to the entire training molecules, it aligns each scaffold with one of the clusters by gathering molecules with the same scaffold into the same cluster that best represents the topological semantics of the scaffold.
In addition, the condition $V\gg K$ induces topologically similar scaffolds to be grouped together.

\smallsection{Remarks}
We also considered two alternative strategies to utilize molecular scaffolds: 
one is minimizing the distance between a molecule's topology representation and its corresponding scaffold representation directly, e.g., $\sum_{v}s_{iv}\cdot||\mathrm{z}_i-\mathrm{e}_v||_2\xspace$, 
and the other is applying scaffold classification loss by using the scaffolds as the label.
We found that the OT-based alignment strategy consistently obtains higher performance than the alternatives (This will be discussed in Appendix B).
Furthermore, the clustering part in \proposed (i.e., GNN, MLP, Eq. \eqref{eq:cluster_loss}, and Eq. \eqref{eq:align_loss}) can be used as a standalone model for deep molecule clustering. 
We name it \textit{topology-based deep clustering}, and further details are provided in Appendix C.

\subsection{Optimization and Inference}
All the parameters in \proposed, including the GNN, the experts, and the gating module, are optimized in an end-to-end manner by minimizing the following loss:
\begin{equation}
\label{eq:loss}
    \mathcal{L}= \mathcal{L}_{class} + \alpha\mathcal{L}_{cluster} +\beta \mathcal{L}_{align},
\end{equation}
where $\alpha$ and $\beta$ are the balance parameters that control the quality of the final topological clusters. 

To make a deterministic prediction, a new input molecule at test time should be assigned into the clusters in a deterministic way.
Thus, it computes the final prediction by using Eq.~\eqref{eq:gating} that replaces $g_{ik}$ with $q_{ik}$. 
In other words, the cluster assignment probability $q_{ik}$ (in Eq.~\eqref{eq:cluster}) is directly utilized as the weight for the experts, without applying Gumbel-Softmax to get $g_{ik}$ (in Eq.~\eqref{eq:gumbel}).

\section{Experiments}

\begin{table*}[t]

\renewcommand{\arraystretch}{0.7}
\fontsize{9}{9}\selectfont
\setlength{\tabcolsep}{5.0pt}

\centering
{
\begin{tabular}{clccccccccc}
    \toprule
    GNN & Method & BBBP & Tox21 & ToxCast & SIDER &ClinTox & MUV & HIV & BACE & AVG.\\\midrule
    & \# of Mol. & 2,039 & 7,831 & 8,575 & 1,427 & 1,478 & 93,087 & 41,127 & 1,513 &- \\
    & \# of Tasks & 1 &12 & 617 & 27 & 2& 17 & 1 & 1 & - \\\midrule 

    \multirow{7}{*}{\rotatebox[origin=c]{90}{GCN}}

    & SingleCLF & 65.9 $\pm$ 0.9  &  74.4 $\pm$ 0.6  &  \textbf{63.6} $\pm$ 1.1  &  60.6 $\pm$ 0.8  &  55.4 $\pm$ 3.6  &  74.0 $\pm$ 1.3  &  75.2 $\pm$ 1.4  &  71.0 $\pm$ 4.6  & 67.5  \\\cmidrule{2-11}
    
    & MoE & 65.4 $\pm$ 1.6 & 74.3 $\pm$0.4  & 61.5 $\pm$ 0.9 & \textbf{60.8} $\pm$ 1.0 & 68.1 $\pm$ 2.4 & 73.9 $\pm$ 1.2 & 75.5 $\pm$ 1.0  & 75.8 $\pm$ 2.8 & 69.4  \\
    
    & E-Ensemble & 65.8 $\pm$ 2.8 & 74.4 $\pm$ 0.6 & 61.6 $\pm$ 0.7 & 60.3 $\pm$0.9  & \textbf{70.5} $\pm$ 5.9 & 74.1 $\pm$ 1.0 & 74.9 $\pm$ 1.0& 76.0 $\pm$ 0.2 & 69.7  \\
    
    & GraphDIVE & 62.5 $\pm$ 1.9 & 73.2 $\pm$ 0.9 & 62.0 $\pm$ 0.7 & 51.9 $\pm$ 2.0 & 60.4 $\pm$ 6.2 & 64.6 $\pm$ 4.7 & 65.2 $\pm$ 4.1 & 59.4 $\pm$ 4.6 & 62.4  \\\cmidrule{2-11}
    
    &\proposed & \textbf{67.0} $\pm$ 2.4 & \textbf{75.3} $\pm$ 0.4 & 63.1 $\pm$ 0.6 & \textbf{60.8} $\pm$ 0.9 & 70.1 $\pm$ 6.0 & \textbf{74.6} $\pm$ 0.6 & \textbf{76.6} $\pm$ 0.7 & \textbf{76.4} $\pm$ 1.9 & \textbf{70.5}  \\
    \midrule
    
    \multirow{6}{*}{\rotatebox[origin=c]{90}{GraphSAGE}}
    & SingleCLF &  \textbf{68.1} $\pm$ 1.5  &  74.2 $\pm$ 0.8  &  \textbf{63.6} $\pm$ 0.7  &  59.7 $\pm$ 1.0  &  53.4 $\pm$ 2.4  &  74.5 $\pm$ 2.5  &  74.6 $\pm$ 1.5  & 70.8 $\pm$ 3.3  & 67.4   \\\cmidrule{2-11}
    
    & MoE &  66.9 $\pm$ 2.0  &  74.5 $\pm$ 0.4  &  62.9 $\pm$ 0.7  &  61.7 $\pm$ 1.3  &  60.3 $\pm$ 4.4  &  73.0 $\pm$ 1.6  &  73.5 $\pm$ 1.0 &  71.1 $\pm$ 3.1 & 68.0  \\
    
    & E-Ensemble & 67.3 $\pm$ 1.6  &  74.5 $\pm$ 0.5  &  62.4 $\pm$ 0.6  &  59.6 $\pm$ 0.8  & \textbf{60.9} $\pm$ 3.2  & 73.6 $\pm$ 6.1  & 75.0 $\pm$ 1.1 & 70.1 $\pm$ 2.5  & 67.9  \\
    
    & GraphDIVE & 61.3 $\pm$ 2.3 & \textbf{74.6} $\pm$ 0.5  & 62.3 $\pm$ 0.6  & 57.1 $\pm$ 2.8  & 57.1 $\pm$ 7.6  & 68.2 $\pm$ 3.9  &  65.2 $\pm$ 2.8  & 65.9 $\pm$ 5.2  & 64.0  \\\cmidrule{2-11}
    
    &\proposed &  67.6 $\pm$ 2.0  &  74.3 $\pm$ 0.5  & 62.6 $\pm$ 0.7  & \textbf{62.6} $\pm$ 0.9  & 58.7 $\pm$ 2.7  & \textbf{76.0} $\pm$ 1.5  & \textbf{75.5} $\pm$ 1.0 &  \textbf{74.0} $\pm$ 3.1  & \textbf{68.9 } \\\midrule

    \multirow{7}{*}{\rotatebox[origin=c]{90}{GAT}}
    
    &SingleCLF             & 64.9 $\pm$ 1.2  & \textbf{75.0}   $\pm$ 0.8  & \textbf{63.5} $\pm$ 1.6  & 61.0   $\pm$ 1.1  & 58.9 $\pm$ 1.4  & \textbf{74.5} $\pm$ 0.9  & 75.5 $\pm$ 1.7  & 75.3 $\pm$ 2.4  & 68.6 \\\cmidrule{2-11}
    
    &MoE & 64.0   $\pm$ 1.4  & 70.8 $\pm$ 0.7  & 62.3 $\pm$ 0.8  & 60.0   $\pm$ 1.6  & 54.1 $\pm$ 4.7  & 73.1 $\pm$ 1.8  & 73.0   $\pm$ 1.9  & \textbf{76.5} $\pm$ 2.8  & 66.7 \\
    
    &E-Ensemble    & \textbf{66.8} $\pm$ 1.5  & 72.2 $\pm$ 1.2  & 62.5 $\pm$ 0.6  & 59.4 $\pm$ 4.1  & 58.7 $\pm$ 4.4  & 73.5 $\pm$ 1.5  & 75.2 $\pm$ 1.1  & 77.0   $\pm$ 3.2  & 68.2 \\
    
    &GraphDIVE          & 64.1 $\pm$ 1.4 & 70.1 $\pm$ 1.3 & 60.4 $\pm$ 1.3 & 53.7 $\pm$ 1.7  & \textbf{60.2} $\pm$ 7.2  & 73.1 $\pm$ 1.6 & 75.5 $\pm$ 1.4 & 68.4 $\pm$ 7.5 & 65.7 \\\cmidrule{2-11}
    
    &\proposed          & 65.4 $\pm$ 2.1 & 74.9 $\pm$ 0.8 & 62.9 $\pm$ 0.9 & \textbf{62.0}   $\pm$ 1.3 & 59.1 $\pm$ 2.5 & 74.1 $\pm$ 1.1 & \textbf{77.3} $\pm$ 1.3 & 76.3 $\pm$ 2.0 & \textbf{69.0}   \\\midrule

    \multirow{7}{*}{\rotatebox[origin=c]{90}{GIN}}
    &SingleCLF             & 68.9  $\pm$ 1.2 & 74.3  $\pm$ 0.6  & \textbf{64.1}  $\pm$ 1.6  & 58.1  $\pm$ 1.5  & 58.8  $\pm$ 5.7  & \textbf{76.1}  $\pm$ 1.3 & 75.6  $\pm$ 1.6  & 69.0    $\pm$ 4.7  & 68.1 \\\cmidrule{2-11}
    
    &MoE & 66.3  $\pm$ 2.0    & 74.5  $\pm$ 0.5  & 60.1  $\pm$ 1.0    & 58.6  $\pm$ 0.9  & 55.5  $\pm$ 3.0    & \textbf{76.1}  $\pm$ 0.8  & 71.4  $\pm$ 2.7  & 68.8  $\pm$ 3.9  & 66.4 \\
    
    &E-Ensemble    & 66.5  $\pm$ 2.0    & 74.4  $\pm$ 1.1  & 60.7  $\pm$ 1.1  & 56.1  $\pm$ 1.6  & 59.8  $\pm$ 7.2  & 72.8  $\pm$ 2.5  & 76.2  $\pm$ 1.1  & 68.3  $\pm$ 5.2  & 66.9 \\
    
    &GraphDIVE          & 65.0    $\pm$ 2.6 & 72.1  $\pm$ 3.0    & 54.7  $\pm$ 1.2  & 52.9  $\pm$ 2.3 & 52.9  $\pm$ 6.9 & 65.5  $\pm$ 7.0 & 68.9  $\pm$ 2.1 & 62.5  $\pm$ 4.7 & 61.8 \\\cmidrule{2-11}
    
    &\proposed          & \textbf{70.0}    $\pm$ 0.7 & \textbf{75.3}  $\pm$ 0.7 & 62.6  $\pm$ 0.4 & \textbf{58.9}  $\pm$ 1.2 & \textbf{60.3}  $\pm$ 4.5 & 75.7  $\pm$ 1.6 & \textbf{76.3}  $\pm$ 1.4 & \textbf{71.7}  $\pm$ 4.0 & \textbf{69.1} \\\midrule

    \multirow{7}{*}{\rotatebox[origin=c]{90}{\shortstack{Pre-trained\\GIN}}}

    &SingleCLF  & 68.7 $\pm$ 1.3  & 78.1 $\pm$ 0.6  & 65.7 $\pm$ 0.6  & \textbf{67.7} $\pm$ 0.8  & 72.6 $\pm$ 1.5  & 81.3 $\pm$ 2.1 & 79.9 $\pm$ 0.7  & 84.5 $\pm$ 0.7  & 74.2\\\cmidrule{2-11}
    
    &MoE & 67.2 $\pm$ 2.3  & 78.3 $\pm$ 0.5  & 65.4 $\pm$ 0.4  & 61.9 $\pm$ 0.9  & 73.2 $\pm$ 2.2  & 80.8 $\pm$ 1.0    & 80.7 $\pm$ 0.5  & 84.7 $\pm$ 0.4 & 74.0   \\
    
    &E-Ensemble    & 67.5 $\pm$ 2.8  & 78.3 $\pm$ 0.3  & 65.9 $\pm$ 0.2  & 62.3 $\pm$ 0.6  & 71.6 $\pm$ 1.0    & 80.4 $\pm$ 1.0    & 80.6 $\pm$ 0.5  & 84.3 $\pm$ 0.8  & 73.9 \\
    
    &GraphDIVE          & 69.1 $\pm$ 0.7 & 78.3 $\pm$ 0.3 & 66.0   $\pm$ 0.3  & 62.4 $\pm$ 0.7 & 72.1 $\pm$ 3.9  & \textbf{82.6} $\pm$ 1.2 & 80.1 $\pm$ 0.5  &  77.8 $\pm$ 2.8 & 73.2 \\\cmidrule{2-11}
    
    &\proposed          & \textbf{69.2} $\pm$ 2.5 & \textbf{78.6} $\pm$ 0.3 & \textbf{66.2} $\pm$ 0.5 & 62.9 $\pm$ 0.7 & \textbf{74.1} $\pm$ 2.1 & 80.3 $\pm$ 0.7 & \textbf{80.8} $\pm$ 0.6 & \textbf{84.9} $\pm$ 0.8 & \textbf{74.9}\\\bottomrule

\end{tabular}}
\caption{Macro ROC-AUC on molecular property prediction tasks. The average score of 8 datasets is reported in the rightmost column. The best score for each dataset under each backbone GNN architecture is marked in bold face.}
\label{tbl:main}
\end{table*}

In this section, we focus on the following three research questions to validate the effectiveness of \proposed.
\begin{itemize}
    \item \textbf{RQ1} Does \proposed effectively predict molecular properties by leveraging the underlying topological patterns?
    \item \textbf{RQ2} Does \proposed provide robust predictions for unseen molecules by capturing topological patterns?
    \item \textbf{RQ3} Do $\mathcal{L}_{cluster}$ and $\mathcal{L}_{align}$ provide useful supervision with the gating module during training?
\end{itemize}
\subsection{Experimental Settings}

\smallsection{Dataset} 
We use 8 benchmark datasets \cite{wu2018moleculenet} for molecular property prediction, and the statistics of the datasets are summarized in Table \ref{tbl:main}.
Following the previous study \cite{hu2020strategies}, we extract molecular features of nodes, edges, and scaffold indices through RDKit.\footnote{http://www.rdkit.org}
For each dataset, we follow the scaffold splitting protocol \cite{hu2020strategies}, which sorts all the molecules by their scaffold indices and then splits them into training/validation/testing sets with a ratio of 80:10:10, respectively. 

\smallsection{Baselines} 
We compare \proposed with various baselines. 
All compared methods use GNN as a feature extractor, but they employ different strategies for making predictions.
\begin{itemize}
    \item \textbf{SingleCLF} is the standard classifier that uses a classification module (i.e., a single expert).
\end{itemize}
The methods below employ multiple experts but use different gating schemes to consolidate the experts.
\begin{itemize}
    \item \textbf{Mixture of Experts (MoE)} employs MLP with Gumbel-Softmax to select relevant experts for each molecule.
    
    \item \textbf{Expert-Ensemble (E-Ensemble)} \cite{dietterich2000ensemble} uses the arithmetic mean to combine the outputs from experts.
    
    \item \textbf{GraphDIVE} \cite{hu2021graph} uses a weighted sum of the outputs from the experts.
    The weights are computed by a linear layer with Softmax. 
\end{itemize}
To evaluate the effectiveness of \proposed with a broad range of GNN backbones \cite{yang2021rethinking}, we use various GNN architectures: \textbf{GCN} \cite{kipf2017semi}, \textbf{GraphSAGE} \cite{hamilton2017inductive}, \textbf{GAT} \cite{velivckovic2017graph}, \textbf{GIN} \cite{xu2018powerful}, and \textbf{Pre-trained GIN} \cite{hu2020strategies}.
For all experiments, we employ 5-layer GNNs and each expert of a single FC layer.

\smallsection{Training Details and Hyper-parameters}
Each model is trained at most 200 epochs, and the training process is terminated when the validation ROC-AUC does not increase for 50 successive epochs.
We train all models ten times with different random seeds and report the average score with its standard deviation.
We search for the best hyper-parameter configuration through a grid search based on the validation ROC-AUC.
The number of experts is chosen from $K\in\{3,5,7,10\}$, and the loss balancing parameters are selected from $\alpha, \beta \in \{5, 1, 0.1, 0.01\}$.
For the temperature annealing of Gumbel-Softmax, the initial temperature $T_0$ is set to 10, and the final temperature $T_E$ is chosen from $\{0.01, 0.1, 1\}$.
Further details are provided in Appendix A.

\subsection{Quantitative Analysis (RQ1)}

Table \ref{tbl:main} reports ROC-AUC scores of each model on 8 benchmark datasets.
The total average scores across 8 datasets indicate the overall capability of predicting molecular properties.
We observe that \proposed consistently shows higher performance than other baselines regardless of its backbone GNN architectures.
Specifically, it achieves the highest average accuracy for all backbone GNNs, yielding higher accuracy than each of the baselines for 35 out of 40 cases on average.
Further, its superior performance to the other multiple experts-based methods shows the effectiveness of our gating module that reflects enriched topological information.
In the case of the Toxcast dataset with 617 tasks, which makes the target prediction much more challenging, \proposed shows marginal improvement or even negative effects over SingeCLF;
however, its performance drop is less severe than the other multiple experts-based methods. 
In particular, GraphDive with a single gate does not perform well in datasets with multiple tasks, as discussed in Hu et al. 2021.

\begin{figure}[t]
    \centering
    \includegraphics[width=\linewidth]{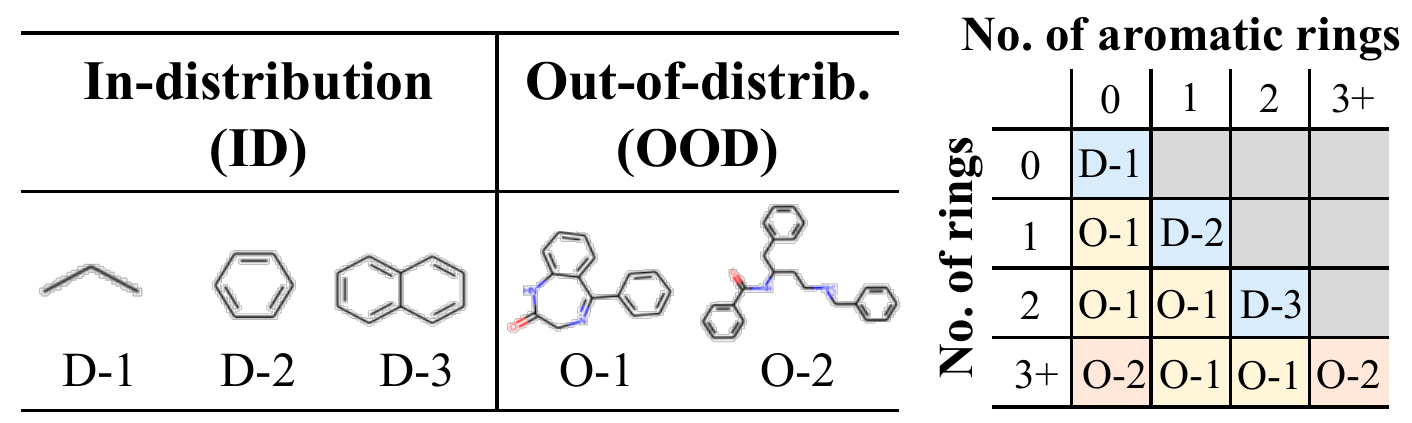}
    \caption{
    Evaluation setup for robustness analysis. 
    Example molecules of each group (Left). 
    Splitting criteria based on the number of rings and aromatic rings (Right).}
    \label{fig:data_split}
\end{figure}

\begin{table}[t]

\centering
\fontsize{9}{9}\selectfont
\setlength{\tabcolsep}{5.0pt}
{
\begin{tabular}{ll|c|cc|cc}
    \toprule
    \multirow{2}{*}{} & \multirow{2.5}{*}{Test data} & \multicolumn{1}{c|}{SingleCLF} & \multicolumn{2}{c|}{Expert-explicit} &  \multicolumn{2}{c}{\proposed} \\\cmidrule{3-7}
    &  &  ROC. & ROC. & Gain  & ROC. & Gain    \\
    \midrule
    
    \multirow{3}{*}{\rotatebox[origin=c]{90}{Tox21}}
    & ID (D-1/2/3) & 77.3 & 69.8 & -7.5 &77.5 &+0.2  \\
    & OOD (O-1) & 68.1 &68.1 &0.0&68.8 &+0.7 \\
    & OOD (O-2) &63.9&65.5&+1.6&64.8&+0.9\\
    \midrule
    
    \multirow{3}{*}{\rotatebox[origin=c]{90}{HIV}}
    & ID (D-1/2/3) & 67.1 & 60.6 & -6.5 &67.2 &+0.1  \\
    & OOD (O-1) & 58.3 & 63.9 & +5.6 & 63.1 & +4.8\\
    & OOD (O-2) & 63.9 & 61.2 & -2.7 & 65.1 & +1.2 \\
    \bottomrule
\end{tabular}
}
\caption{Macro ROC-AUC on ID and OOD groups (Backbone: GIN). Gain denotes the improvement over SingleCLF.}
\label{tbl:general}
\end{table}

\begin{figure}[t]
    \centering
    \includegraphics[width=\linewidth]{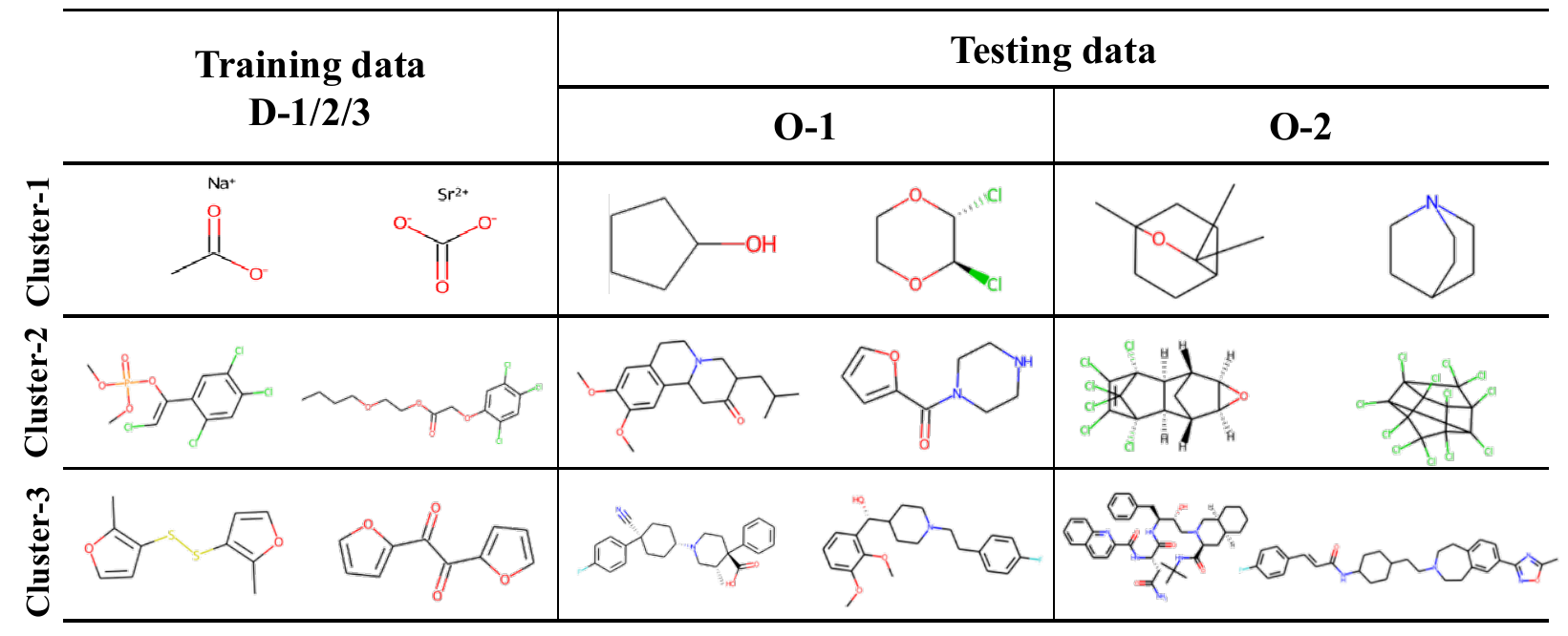}
  \caption{The molecule closest to the three cluster centroids for each molecule group, identified by \proposed.}
    \label{fig:mol_examples}
\end{figure}

\subsection{Robustness Analysis (RQ1 \& RQ2)}
Although the scaffold splitting protocol ensures that there is no common scaffold between the training and testing set, it does not consider high-level topological semantics beyond the scaffolds.
For a more thorough evaluation of the robustness of new molecular structures having topological semantics distinguishable from the training set, we introduce a new splitting protocol. 
Specifically, we divide molecules into five groups, each of which has a disjoint scaffold set, according to the combination of the number of rings and aromatic rings\footnote{It is known that the number of rings is an important descriptor of molecular properties \cite{debnath1991structure}.} (Figure \ref{fig:data_split}). 
We categorize D-1, D-2, and D-3 as in-distribution (ID) and O-1 and O-2 as out-of-distribution (OOD).
Then, we train models using ID molecules only and evaluate the performance using both ID and OOD molecules.
To further ascertain the effectiveness of our gating module, we compare \proposed with \textbf{Expert-explicit}, a variant that explicitly and deterministically assigns an expert for each ID group (i.e., D-1, D-2, and D-3).\footnote{To identify the most relevant expert to each OOD molecule, Expert-explicit utilizes the centroid of each ID group.
We set the number of experts to 3 for both Expert-explicit and \proposed.}

Table \ref{tbl:general} shows the prediction performances on ID and OOD groups.
Compared to SingleCLF, \proposed achieves comparable performance for ID groups and consistently higher performance for OOD groups, strongly supporting the \proposed's robustness to unseen molecular structures.
However, Expert-explicit shows a significant performance drop for ID groups and degraded performance for an OOD group on the HIV dataset.
That is, the molecule segregation based on pre-defined rules leads to suboptimal results because it cannot capture information useful for molecular property prediction.
In contrast, \proposed clusters molecules in favor of predicting their molecular property during optimization, eventually assigning molecules to the experts in a data-driven and task-specific way, which is useful in real-world applications.
In practice, the assignment ratio of training molecule groups (D-1/2/3) to each cluster (Appendix D) varies depending on the dataset, indicating that \proposed adjusts how much it incorporates topology information by itself.    
Figure \ref{fig:mol_examples} also illustrates that \proposed assigns molecules with unseen structures (i.e., O-1 and O-2) to the most topologically similar clusters.

\begin{table}[t]

\centering
{
\fontsize{9}{9}\selectfont
\setlength{\tabcolsep}{4.0pt}

\begin{tabular}{cc|cc|cc|cc}
    \toprule
     \multirow{2}{*}{$\mathcal{L}_{cluster}$} & \multirow{2}{*}{$\mathcal{L}_{align}$} & \multicolumn{2}{c|}{Tox21} & \multicolumn{2}{c|}{HIV} &  \multicolumn{2}{c}{BACE} \\\cmidrule{3-8}
      &  & NMI & ROC. & NMI & ROC. & NMI & ROC.   \\
     \midrule
      & & 0.096 & 74.5 & 0.149 &71.4 & 0.322 & 68.8\\
      $\checkmark$ & & 0.223 & 75.1 &0.347 &75.7 & 0.372 &71.0\\
      & $\checkmark$ & 0.098 &74.6 &0.154  & 72.5 &0.335 &68.7\\

     $\checkmark$ & $\checkmark$ & 0.228 &\textbf{75.3} &0.357 &\textbf{76.3} &0.383 &\textbf{71.7}\\

    \bottomrule
\end{tabular}
}
\caption{Macro ROC-AUC of ablations (Backbone: GIN).}
\label{tbl:ablation}
\end{table}

\subsection{Ablation Study (RQ3)}
To examine how each component of our model affects the final performance, we conduct an ablation study.
Note that without $\mathcal{L}_{cluster}$ and $\mathcal{L}_{align}$, our method is equivalent to the basic MoE.
Table~\ref{tbl:ablation} shows that the clustering loss helps improve the performances over MoE.
Furthermore, the best performance is achieved by using both $\mathcal{L}_{cluster}$ and $\mathcal{L}_{align}$.
Besides, we report normalized mutual information (NMI) \cite{strehl2002cluster} to evaluate how many molecules with the same scaffold belong to the same cluster.
We observe that both losses indeed improve NMI, but the extent of the growth varies widely across datasets.

\subsection{Further Analysis}
\smallsection{Additional cost for multiple experts}
We compare computational costs of \proposed with that of SingleCLF on BBBP dataset.
All computations required for multiple experts are executed in parallel by GPU processors. 
As a result, \proposed shows only 4\% increase in training cost and 2\% increase in inference cost.
In terms of model size, \proposed additionally uses 1\% of the total parameters of the GIN-based model. 
Taken together, we conclude that the additional cost for multiple experts in \proposed is marginal. 

\smallsection{Parameter analysis}
The effects of the loss balancing parameters (i.e., $\alpha$ and $\beta$) and the number of experts (i.e., $K$) are reported in Appendix E and F, respectively.
In short, their optimal values vary depending on the dataset, and they can be easily tuned by using the validation set.

\section{Conclusion}

\proposed adopts a clustering-based gating module that assigns input molecules into topological groups and optimizes it for effective clustering in two self-supervised ways:
1) cluster cohesion strengthening and 2) cluster-scaffold alignment.  
Our extensive experiments demonstrate that \proposed significantly boosts the performance with various backbone GNN architectures, implying that topology-specific experts can offer complementary information for better generalization on test molecules with new topological patterns.

\section{Acknowledgments}
This work was supported by the IITP grant funded by the MSIT (No.2018-0-00584, 2019-0-01906, 2022-00155958), the NRF grant funded by the MSIT (No.2020R1A2B5B03097210), and the Technology Innovation Program funded by the MOTIE (No.20014926).

\bibliography{aaai23}

\clearpage
\appendix

\section{Supplementary Materials For “Learning Topology-Specific Experts for Molecular Property Prediction”}
\subsection{A. More Details about the Experiments} 
\smallsection{Software and Hardware} We implement all the models using PyTorch 1.8.0 \cite{paszke2019pytorch} and PyTorch-geometric 2.0.4 \cite{fey2019fast}. The experiments are conducted on Ubuntu 20.04 LTS with CUDA 11.4. and on GeForce GTX 1080 Ti GPU 11GB.

\smallsection{Hyper-parameters} We use Adam optimizer, and the hyper-parameters we tune for each dataset are: (1) the learning rate $\in \{0.01, 0.001, 0.0001\}$ ; (2) L2 weight decay $\in \{0, 0.0001, 0.00001\}$; (3) the batch size $\in \{32, 512\}$.

\smallsection{Input features} Following the previous study \cite{hu2020strategies}, we use minimum input features that unambiguously describe the two-dimensional structure of molecules. 
For node features, we use atom number ([1, 118]) and chirality tags (unspecified, tetrahedral cw, tetrahedral ccw, other). 
For edge features, we use bond types (single, double, triple, aromatic) and bond directions (–, endupright, enddownright).

\subsection{B. Alternative Strategies to Utilize Molecular Scaffolds } 
Instead of the scaffold alignment (i.e., $\mathcal{L}_{align}$ in Eq. (\ref{eq:align_loss})), we considered two alternative ways to utilize molecular scaffolds. 

\smallsection{Alternative-1} One possible solution is directly minimizing the distance between a molecule's topology representation $\mathrm{z}_i$ and its corresponding scaffold representation $\mathrm{e}_v$ as follows:
\begin{equation}
\label{eq:aliter1_loss}
    \mathcal{L}_{alter1}={\frac{1}{N}}\sum_{i=1}^{N}\sum_{v=1}^{V} s_{iv}\cdot||\mathrm{z}_i-\mathrm{e}_v||_2,
\end{equation}
where $s_{i}$ is a one-hot distribution of sample $i$ over all the scaffolds in a training set.

\smallsection{Alternative-2} 
We apply scaffold classification loss by using the scaffolds as the label.
Let $\mathrm{\hat{s}}_i=\mathrm{MLP}(\mathrm{h}_{G_i}) \in \mathbb{R}^{V}$ denote the classification logits, where $V$ is the number of scaffolds in a training set. 
Then, we use categorical cross-entropy loss as follows:
\begin{equation}
\label{eq:alter2_loss}
    \mathcal{L}_{alter2}={\frac{1}{N}} \sum_{i=1}^{N} \mathrm{CE}(\mathrm{s}_i, \mathrm{\hat{s}}_i).
\end{equation}

\smallsection{Results} In table \ref{tbl:alter_scaffold}, we observe that the OT-based
alignment strategy consistently obtains higher performance than the alternatives. 
It supports the effectiveness of our gating module that reflects enriched topological information. 
Specifically, Alternative-1 can not directly force representations to become closer to centroids, 
and alternative-2 pushes molecules with different scaffolds away from each other, which can hinder  the clustering process and lead to limited performance.

\begin{table}[ht]
\centering
{
\fontsize{9}{9}\selectfont
\setlength{\tabcolsep}{4.0pt}

\begin{tabular}{c|cc|cc|cc}
    \toprule
      & \multicolumn{2}{c|}{Tox21} & \multicolumn{2}{c|}{HIV} &  \multicolumn{2}{c}{BACE} \\\cmidrule{2-7}
        & NMI & ROC. & NMI & ROC. & NMI & ROC.   \\
     \midrule
       Alternative-1 &  0.017&75.1  & 0.357 & 75.4  & 0.305 & 71.6 \\
       Alternative-2 & 0.001 & 74.5 &0.124 & 68.8 & 0.234 & 65.0\\
       \proposed & 0.228 &\textbf{75.3} &0.357 &\textbf{76.3} &0.383 &\textbf{71.7}\\

    \bottomrule
\end{tabular}
}
\caption{Macro ROC-AUC for alternatives (Backbone: GIN).}
\label{tbl:alter_scaffold}
\end{table}

\subsection{C. Topology-based Deep Clustering } 
The clustering part in \proposed can be used as a standalone model for deep molecule clustering. 
The training procedure for Topology-based Deep Clustering is provided in Algorithm 1.

\begin{algorithm}
\caption{Topology-based Deep Clustering}
\label{alg:algorithm}
\textbf{Input}: A training set $D$ (w/t scaffold index for each sample), \# total epochs $T_{ep}$\\
\textbf{Output}: GNN/ MLP parameters and the cluster centroids
  \begin{algorithmic}[1]
  \STATE Initialize centroids by $K$-means clustering on topology representations from a randomly-initialized model
    \FOR{$t=0,1,...,T_{ep}$}
      \FOR{each batch $\mathcal{B} \in D$}
        \STATE Compute $\mathcal{L}_{cluster}$ (Eq. \ref{eq:cluster_loss})
        \STATE Compute $\mathcal{L}_{align}$ (Eq. \ref{eq:align_loss})
        \STATE Compute $\mathcal{L} = \alpha\mathcal{L}_{cluster} +\beta \mathcal{L}_{align}$
        \STATE Update all parameters by minimizing $\mathcal{L}$
      \ENDFOR
    \ENDFOR
  \end{algorithmic}
\end{algorithm}

\begin{table}[ht]
\centering
{
\fontsize{9}{9}\selectfont
\begin{tabular}{cc|c|c|c}
    \toprule
     
      $\mathcal{L}_{cluster}$ & $\mathcal{L}_{align}$ & Tox21 & HIV  & BACE \\\cmidrule{1-5}
      $\checkmark$ & & 0.3825  & 0.3896  & 0.3963 \\
      & $\checkmark$ & 0.2981 & 0.1083   & 0.0001 \\
     $\checkmark$ & $\checkmark$ & 0.3991  & 0.3980  & 0.3980 \\
    
    \bottomrule
\end{tabular}
}
\caption{NMI of the clustering results (Backbone: GIN).
}
\label{tbl:cluster_abl}
\end{table}

\begin{figure}[ht]
    \centering
    \includegraphics[scale=0.45]{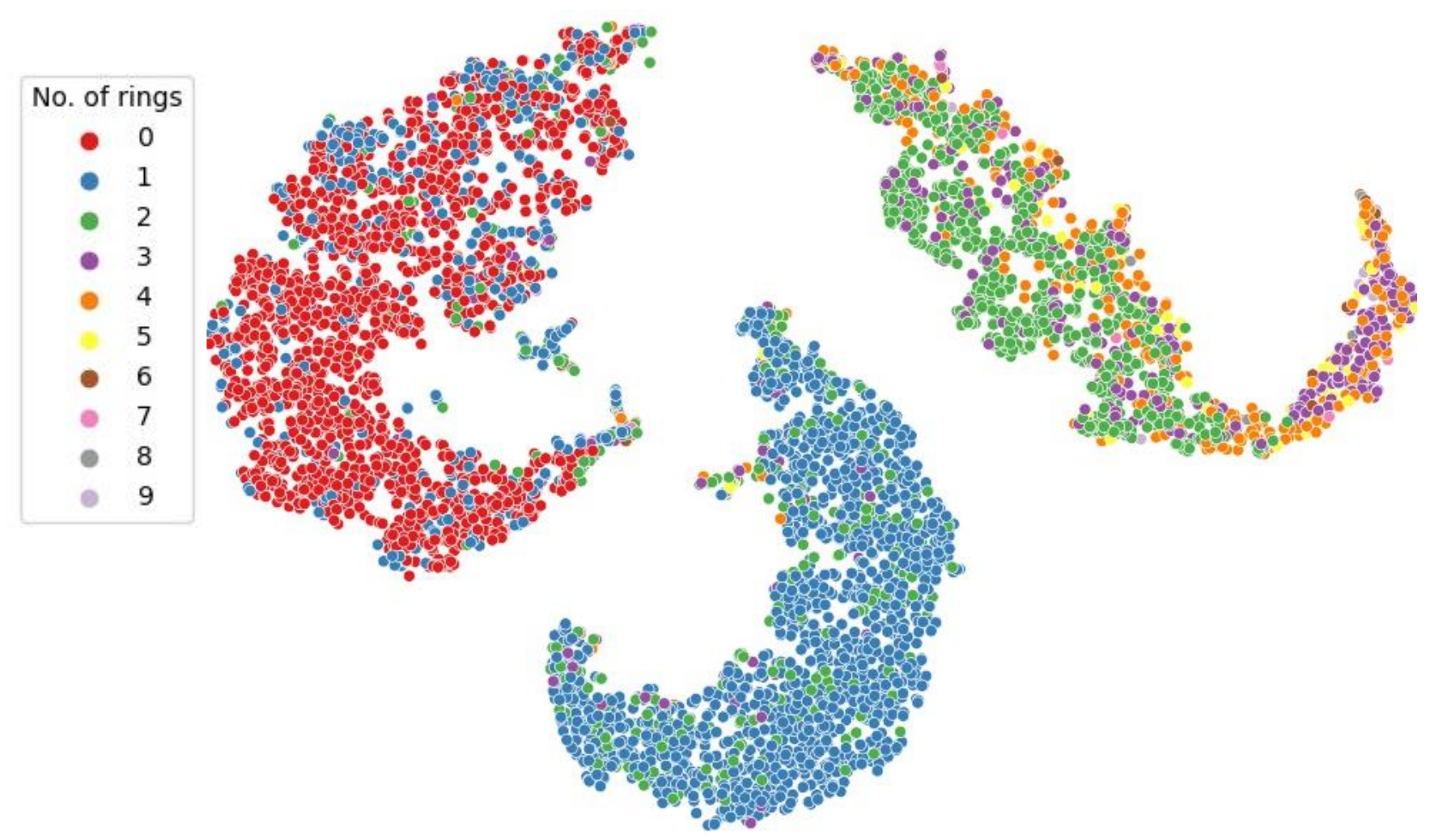}
    \caption{
    The t-SNE visualization of molecule's topology representations on Tox21 using topology-based deep clustering method (Backbone: GIN).}
    \label{fig:tsne_cluster}
\end{figure}

In table \ref{tbl:cluster_abl}, we report NMI of the clustering results. 
We observe that both losses of topology-based deep clustering indeed improve NMI, but the extent of growth varies widely across datasets. 
It is worth noting that the cohesive clustering (Eq. \ref{eq:cluster_loss}) and scaffold alignment (Eq. \ref{eq:align_loss}) collaboratively evolve during the optimization. 
To be specific, the clustering loss strengthens the cohesion of each cluster by sharpening the assignment probability $\mathrm{q}$,
and this further boosts molecules with the same scaffold to be densely gathered to the same cluster.
Further, we plot the t-SNE visualization for Tox21 dataset in Figure \ref{fig:tsne_cluster}. 
We set the number of clusters as 3. 
It shows that molecules having similar structural patterns are densely located together.
We leave applying the topology-based deep clustering to other downstream tasks for future study.

\subsection{D. Robustness Analysis: The assignment ratio of training data groups (D-1/2/3) to each cluster} 

\begin{table}[ht]
\centering
{
\fontsize{9}{9}\selectfont
\setlength{\tabcolsep}{4.0pt}

\begin{tabular}{c|ccc|ccc}
    \toprule
    & \multicolumn{3}{c|}{Tox21} & \multicolumn{3}{c}{HIV} \\ \cmidrule{2-7}
    & D-1 & D-2 & D-3 & D-1 & D-2 & D-3 \\ 
     \midrule
     Cluster-1 & 1  & 0.06 & 0 & 1 & 0.03 & 0  \\
     Cluster-2 & 0  & 0.93 & 0.01 & 0  & 0.62  & 0.12 \\
     Cluster-3 & 0  & 0.01 & 0.99 & 0 & 0.35 & 0.88    \\               
    \bottomrule
\end{tabular}
}
\caption{The assignment ratio of training data groups to each cluster.}
\label{tbl:assign}
\end{table}

In table \ref{tbl:assign}, we report the assignment ratio of the training data to each cluster by \proposed. 
We can see that the cluster assignments found by \proposed are considerably different from the groups divided based on pre-defined rules (i.e., the number of rings).
For instance, a considerable amount of molecules belonging to D-2 and D-3 are assigned to the same cluster on the HIV dataset.
This result again demonstrates that \proposed can cluster molecules in a data-driven and task-specific way tailored to predict their molecular properties.

\subsection{E. Effects of the Loss Balancing Parameters} 
We provide analyses to guide how to select the loss balancing parameters:
$\alpha$ controlling the effects of the clustering loss and $\beta$ controlling the effects of the alignment loss.
For the sake of space, we report the results GCN and GIN on Tox21 dataset. 

\begin{figure}[ht]
	\centering
 
	\begin{subfigure}{0.49\columnwidth}
	    \centering
	    \includegraphics[width=\textwidth]{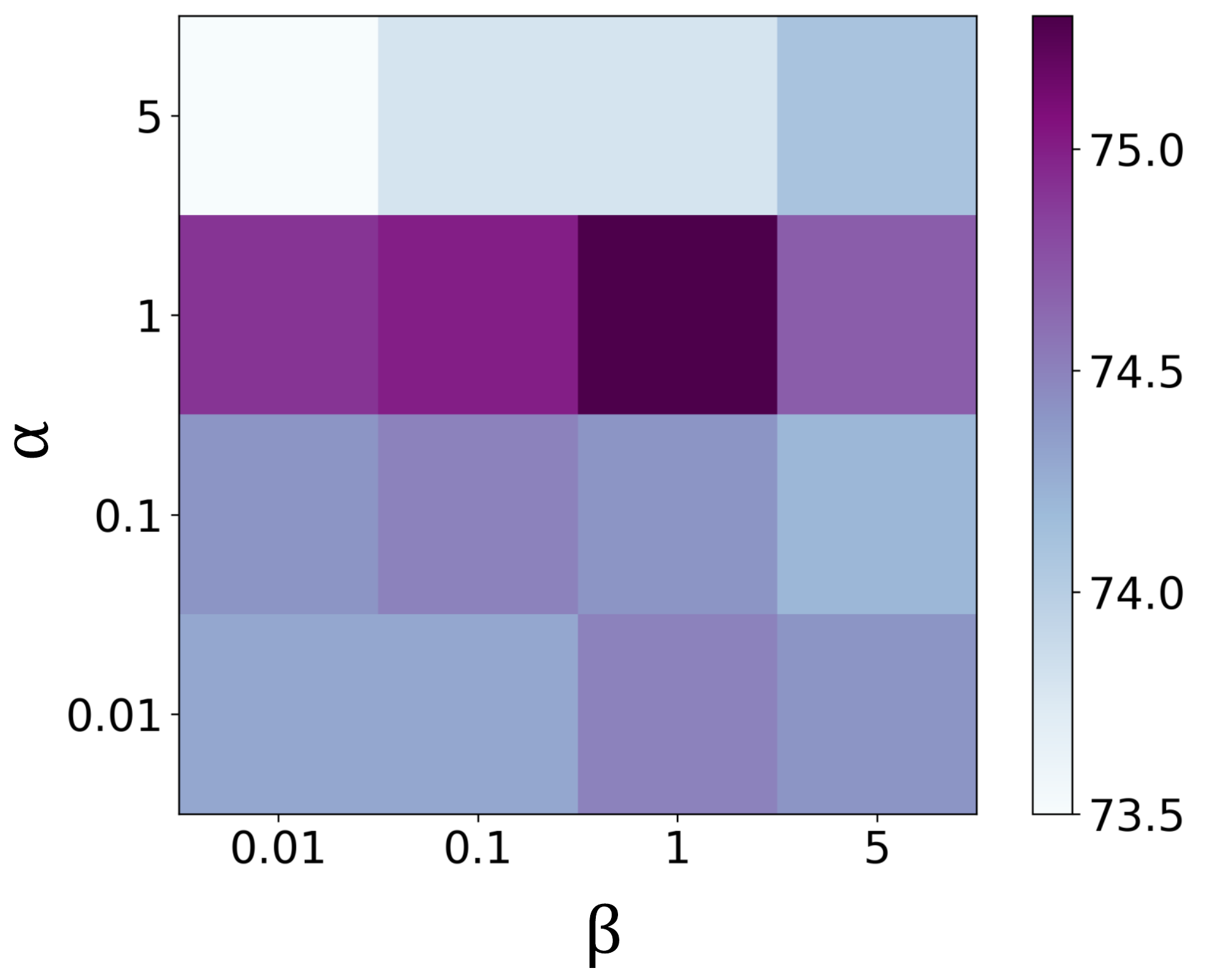}
	    \caption{GCN}
	\end{subfigure}
	\hfill
    \begin{subfigure}{0.49\columnwidth}
	    \centering
	    \includegraphics[width=\textwidth]{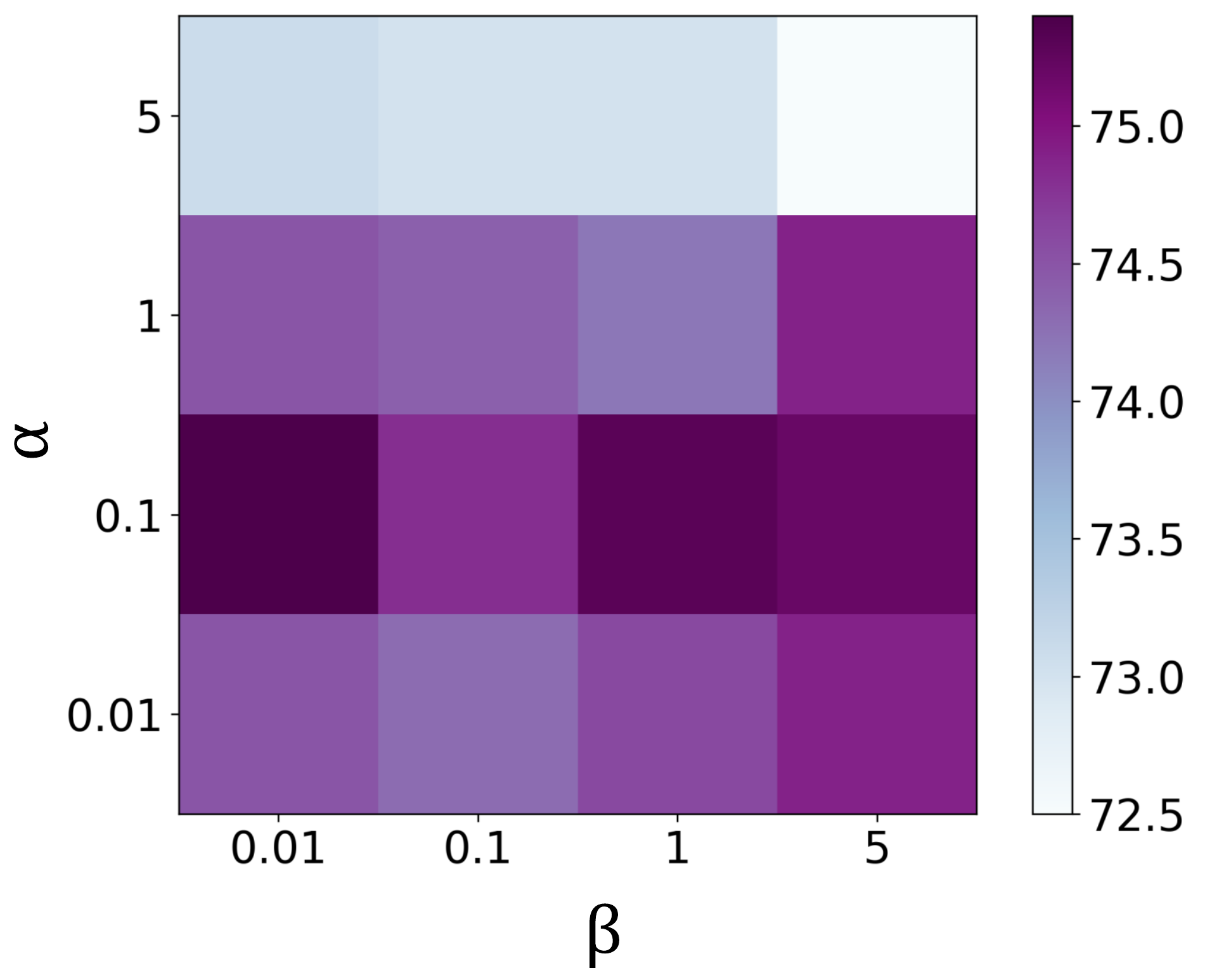}
	    \caption{GIN}
	\end{subfigure}

\caption{Effects of loss balancing parameters. Macro ROC-AUC on Tox21 dataset is reported. }
\label{fig:loss_bal_combination}

\end{figure}

In figure \ref{fig:loss_bal_combination}, $\alpha$ is an important factor affecting the performance of \proposed. 
When $\alpha$ is too large, the clustering objective overwhelms the classification process, leading to limited performance.
Therefore, proper balancing between task-specific and topological clustering objectives is important.

\subsection{F. Effects of the Number of Experts}

\begin{figure}[t]
    \centering
    \includegraphics[width=\linewidth]{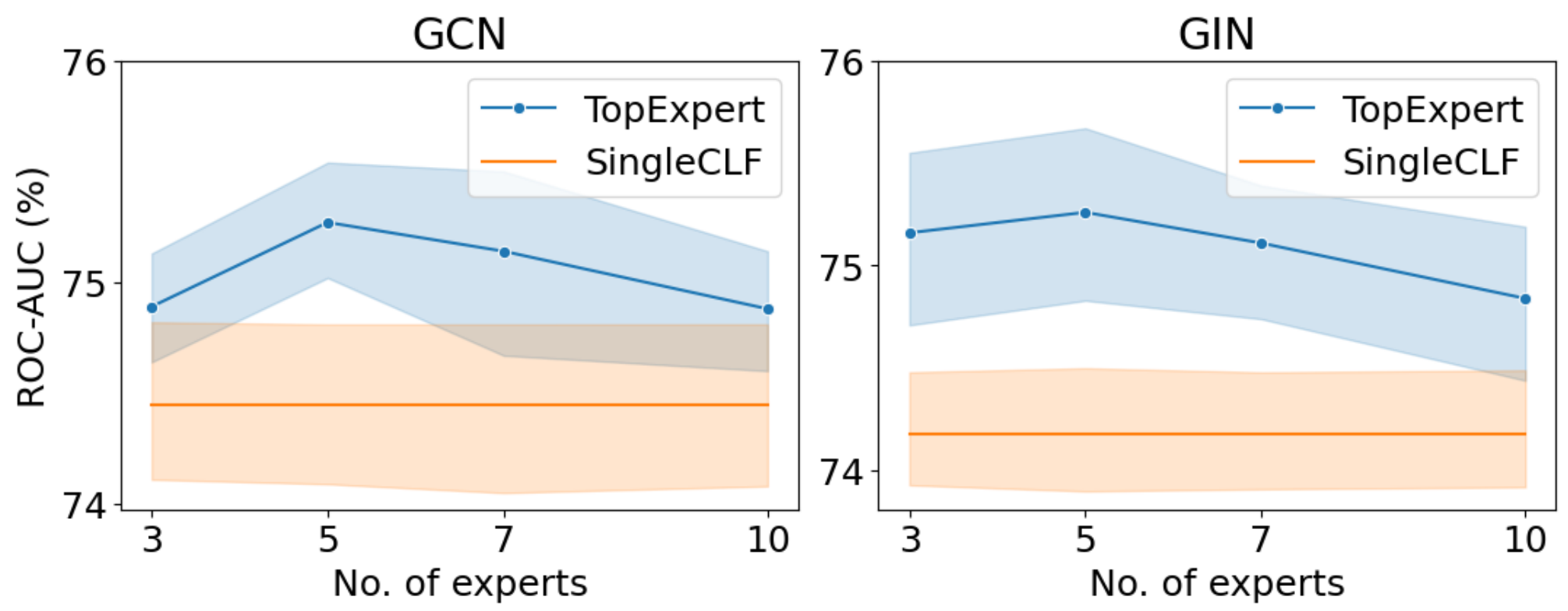}
    \caption{Macro ROC-AUC w.r.t. the number of experts (Dataset: Tox21). The band gap shows 95$\%$ confidence intervals.}
    \label{fig:num_cluster}
\end{figure}

Figure~\ref{fig:num_cluster} shows the performance change of \proposed with respect to the number of experts.
\proposed shows a stable performance around 5-7 and consistently outperforms SingleCLF for all the cases, which again supports the superiority of the proposed framework.
The number of experts needs to be adjusted depending on the dataset, as the optimal varies depending on the dataset.

\end{document}